\documentclass[10pt,twocolumn,letterpaper]{article}

\usepackage{wacv}
\usepackage{times}
\usepackage{epsfig}
\usepackage{graphicx}
\usepackage{amsmath}
\usepackage{amssymb}
\usepackage{multirow}
\usepackage[symbol*]{footmisc}
\usepackage{caption}
\usepackage[export]{adjustbox}
\usepackage{subfig}
\usepackage{booktabs}
\usepackage{float}
\usepackage[dvipsnames]{xcolor}
\usepackage{textcomp}
\usepackage{pbox}

\def\wacvPaperID{****} 

\wacvfinalcopy 

\ifwacvfinal
\def\assignedStartPage{1} 
\fi

\ifwacvfinal
\usepackage[breaklinks=true,bookmarks=false]{hyperref}
\else
\usepackage[pagebackref=true,breaklinks=true,colorlinks,bookmarks=false]{hyperref}
\fi

\ifwacvfinal
\else
\fi

\def\httilde{\mbox{\tt\raisebox{-.5ex}{\symbol{126}}}}

\begin{document}

\title{IQ-VQA: Intelligent Visual Question Answering}

\author{Vatsal Goel\thanks{Equal Contribution} \and Mohit Chandak\footnotemark[1] \and Ashish Anand \and Prithwijit Guha \and Indian Institute of Technology, Guwahati\\
{\tt\small \{vatsal29, mohit, anand.ashish, pguha\}@iitg.ac.in}}

\maketitle

\begin{abstract}
  Even though there has been tremendous progress in the field of Visual Question Answering, models today still tend to be inconsistent and brittle. To this end, we propose a model-independent cyclic framework which increases consistency and robustness of any VQA architecture. We train our models to answer the original question, generate an implication based on the answer and then also learn to answer the generated implication correctly. As a part of the cyclic framework, we propose a novel implication generator which can generate implied questions from any question-answer pair. As a baseline for future works on consistency, we provide a new human annotated VQA-Implications dataset. The dataset consists of \httilde{30k} questions containing implications of 3 types - Logical Equivalence, Necessary Condition and Mutual Exclusion - made from the VQA v2.0 validation dataset. We show that our framework improves consistency of VQA models by \httilde{15\%} on the rule-based dataset, \httilde{7\%} on VQA-Implications dataset and robustness by \httilde{2\%}, without degrading their performance. In addition, we also quantitatively show improvement in attention maps which highlights better multi-modal understanding of vision and language.
\end{abstract}

\section{Introduction} \label{sec:intro}

Visual Question Answering \cite{vqa} task requires an AI system to answer natural language questions on a contextual image. Ideally, this system should be equipped with the ability to extract useful information (with reference to the question) by looking at the image. To answer these questions correctly, the system should not only identify the color, size, or shape of objects, but may also require general knowledge and reasoning abilities which makes the task more challenging. 

Previous works \cite{biasdis,vqa2} have pointed out strong language priors present in the VQA dataset. This could result in false impression of good performance by many state of the art models, without them actually understanding the image. For instance, answering any question starting with ``What sport is" by ``tennis" results in 41\% accuracy. Moreover, citing the `visual priming bias' present in the VQA dataset, questions starting with ``Do you see a .." result in "yes" 87\% of the time. 

\begin{figure}
    \centering
    \subfloat[Input image]{
        \includegraphics[scale=0.21]{images/cons.jpg}
        \label{fig:ex_img}
    } \\
    \subfloat[Implications and Rephrasings answered incorrectly]{
    	\begin{tabular}{lrr}
		\toprule
		\textbf{Original} & How many sailboats are there? & \textcolor{ForestGreen}{1}\\
		\midrule
		\textbf{Logeq} & Is there 1 sailboat? & \textcolor{red}{no}\\
		\textbf{Mutex} & Are there 2 sailboats? & \textcolor{red}{yes}\\
		\textbf{Nec} & Are there any sailboats? & \textcolor{red}{no}\\
		\textbf{Rep} & What is the number of sailboats? & \textcolor{red}{2}\\
		\textbf{Rep} & How many sailboats can you see? & \textcolor{red}{2}\\
		\textbf{Rep} & How many sailboats do you see? & \textcolor{red}{2}\\
		\bottomrule
		\end{tabular}
    }
    \caption{\textbf{An example of inconsistent and brittle nature of VQA models}. Even though the model \cite{pythia} correctly answers the original question, it fails to answer any of the 3 generated implications and rephrasings correctly.}
    \label{fig:cons}
\end{figure}

Many recent works \cite{imps,cycle,lol} have shown that despite having high accuracy on questions present in the dataset, these models perform poorly when rephrased or implied questions are asked and hence are not intelligent enough to be deployed in the real world. Fig \ref{fig:cons} shows the inconsistent and brittle nature of VQA models. Despite answering the original question correctly, the model fails to answer rephrased and implied questions related to the original question. This shows that models learn from language biases in the dataset to some extent, rather than correctly understanding the context of the image.

\begin{figure}[t]
    \centering
    \subfloat[Input image example]{
        \includegraphics[scale=0.5]{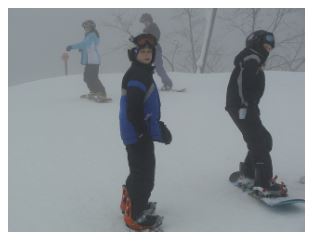}
        \label{fig:imp_img}
    } \\
    \subfloat[Generated implication example]{
    	\begin{tabular}{lrr}
		\toprule
		\textbf{Original} & How many people? & 4 \\
		\midrule
		\textbf{Logeq} & Are there 4 people? & yes \\
		\textbf{Mutex} & Are there 5 people? & no \\
		\textbf{Nec} & Are there any people? & yes \\
		\bottomrule
		\end{tabular}
    }
    \caption{\textbf{An example of the rule-based implication dataset}. Note that the implications can be answered without looking at the image.}
    \label{fig:imp_ex}
\end{figure}

Throughout the paper, \textbf{Implications} are defined as questions $Q_{imp}$ which can be answered by knowing the original question $Q$ and answer $A$ without the knowledge of the context i.e. image $I$. We categorize these implications into 3 types - logical equivalence, necessary condition and mutual exclusion - as introduced in \cite{imps}. Fig \ref{fig:imp_ex} shows these 3 categories for a QA pair. \textbf{Consistency} is the percentage of implications answered correctly, given that the original question is answered correctly. \textbf{Rephrasings} $Q_R$ are linguistic variations on original question $Q$ keeping the answer $A$ exactly same, as introduced in \cite{cycle}. \textbf{Robustness} is defined as the accuracy on rephrasings, calculated only on correctly answered original questions.

We believe that any model can be taught to better understand the content of the image by enforcing intelligence through consistency and robustness among the predicted answers. In this paper, we present and demonstrate a cyclic training scheme to solve the above mentioned problem of intelligence. Our framework is model independent and can be integrated with any VQA architecture. The framework consists of a generic VQA module and our implication generation module tailored especially for this task.

Our framework ensures intelligent behaviour of VQA models while answering different questions on the same image. This is achieved in two steps: Implication generator module introduces linguistic variations in the original question based on the answer predicted by the VQA model. Then, the model is again asked to answer this on-the-fly generated question so that it remains consistent with the previously predicted answer. Thus, the VQA architecture is collectively trained to answer questions and their implications correctly. We calculate the consistency of different state of the art models and show that our framework significantly improves consistency and robustness without harming the performance of the VQA model.

We observe that there is no benchmark for consistency, which perhaps is the reason for limited development in this area. Hence, to promote robust and consistent VQA models in the future we collect a human annotated dataset of around 30k questions on the original VQA v2.0 validation dataset.\footnote{Dataset and code will be made publicly available.}

In later sections, we demonstrate the quality of these generated questions. We provide a baseline of our implication generator module for future works to compare with. We also perform a comparative study of the attention maps of models trained with our framework to those of baselines. We provide a qualitative and quantitative analysis and observe significant improvement in the quality of these attention maps. This proves that by learning on these variations, our framework not only improves the consistency and robustness of any generic VQA model but also achieves a stronger multi modal understanding of vision and language.

To summarize, our main contributions in this paper are as follows -
\begin{itemize}
    \item We propose a model independent cyclic framework which improves consistency and robustness of any given VQA architecture without degrading the architecture's original validation accuracy.
    \item We propose a novel implication generator module, which can generate implications $G : (Q,A) \longrightarrow Q_{imp}$, for any given question answer pair.
    \item For future evaluation of consistency, we provide a new VQA-Implication dataset. The dataset consists of \httilde{30k} questions containing implications of 3 types - Logical Equivalence, Necessary Condition and Mutual Exclusion.
\end{itemize}

\section{Related Works}

Ever since Visual Question Answering \cite{vqa} was introduced, numerous models have been proposed to combine techniques from computer vision and natural language processing using techniques such as CNNs and LSTMs. Some of the best models using complex attention mechanisms include \cite{butd,BAN,pythia}. The current state of the art is LXMERT \cite{lxmert}, which uses a transformer network for self and cross attention between vision and language modalities.

Analysis of the VQA v1 dataset \cite{vqa2,vqa3} showed the presence of language priors in the dataset. Models were reportedly exploiting these priors as a shortcut for answering questions instead of understanding the image. To tackle this problem, VQA 2.0 was released which created complementary pairs in order to counter these priors. More specifically, for every image, question and answer triplet $(I,Q,A)$, a complimentary image $I^c$ and answer $A^c$ were created. However, investigations in \cite{biasdis} found that even after these ramifications priors continue to exist and exploited by VQA models.

Recent works \cite{cycle,imps,lol} in VQA have introduced novel benchmarks such as robustness and consistency of models as a step towards limiting the false sense of progress in VQA with just accuracy and proposing models with better multi-modal understanding.

\textbf{Consistency:} Inconsistency in QA models on the VQA and SQUAD dataset was first studied in \cite{imps}. They show how even the best VQA models are inconsistent in their answers. For example, given a question "How many birds are in the picture?", the model correctly answers "3". But upon asking "Are there 3 birds in the picture?", the same model incorrectly answers "No". This shows that models lack high level language and vision capabilities and could still be exploiting biases present in the dataset. They proposed evaluating consistency of these models and a simple data augmenting technique for improvement. However, augmentation limits the scope of implied questions to the added dataset. We in-turn propose a generative model based solution without this limitation.

More recent work in \cite{lol} tackles inconsistency among binary i.e. "yes/no" questions. They argue that despite answering original question correctly, VQA models performs poorly on logical composition of these questions. Another work \cite{squint}, focuses on improving consistency of models on reasoning questions. Unlike \cite{imps}, these works provide model based solution but they target only a specific category of questions such as reasoning or binary questions. Unlike these, we show that our approach works better on the entire VQA v2.0 dataset rather than a small subset of it.

Authors of \cite{convqa} previously attempted to improve consistency on the entire VQA v2.0 dataset. However, their concept of Entailed questions, generated from the Visual Genome \cite{vg} dataset, is quite different to our Implications. We believe that if the model is able to answer a question correctly, it should also be able to answer its implications correctly, which implies consistent behavior. But in \cite{convqa}, given a question $Q$ as "Has he worn sneaker?" and answer "yes", an entailed question $Q'$ is "Where is this photo?" with answer "street". Clearly $Q$ and $Q'$ have no direct relation and as per our definition of consistency, answering $Q$ and $Q'$ correctly does not exhibit consistent behavior.

\textbf{Robustness:} To decrease strong language priors, a number of works \cite{vqa2,vqa4} have introduced balanced datasets in context of robustness. The concept of robustness as a measure of performance on linguistic variations in the questions known as rephrasings was first introduced by \cite{cycle}. However, they used a 'consensus score' metric whereas we use a metric similar to consistency for evaluation, to provide uniformity. To motivate future works in this field, they provide a VQA-Rephrasings dataset which we use to evaluate robustness of our models. 

\begin{figure*}[t]
     \centering
     \subfloat[]{
         \centering
         \includegraphics[width=0.69\textwidth,valign=c]{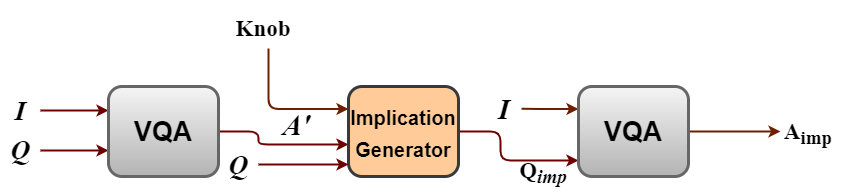}
         \vphantom{\includegraphics[width=0.27\textwidth,valign=c]{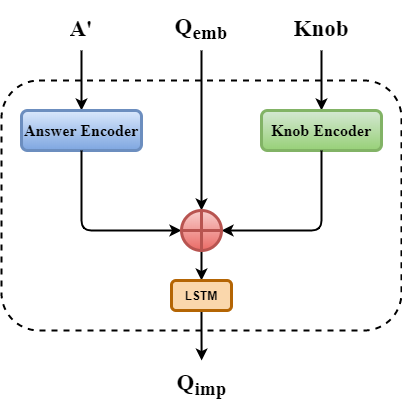}}
         \label{fig:model_diag}}
     \hfill
     \vline
     \hfill
    \subfloat[]{
         \centering
         \includegraphics[width=0.27\textwidth,valign=c]{images/qg_diag.png}
         \label{fig:qg_diag}}
     \label{fig:model}
     \caption{\textbf{ Proposed Model Architecture} (a) Abstract representation of our cyclic framework. Given an input image $I$ and question $Q$, a VQA model predicts answer $A'$. Then our proposed Implication generator transforms the original question $Q$ to $Q_{imp}$ using $A'$ and a control knob. This generated implication (and image) is passed to the VQA model to obtain answer $A_{imp}$. (b) Detailed architecture of our implication generator. The predicted answer $A'$ and control knob are encoded to a latent space using respective encoders. They are then summed up along with question embedding and fed to a LSTM to generate implication $Q_{imp}$.}
\end{figure*}

\textbf{Question Generation:} There has been a thorough study of Natural Language generation in NLP, such as \cite{qg_1,qg_2,qg_3,qg_4}. \cite{qg_2} extracts keywords from knowledge graphs and then formulate question generation from these keywords as Seq2Seq translation problem. \cite{qg_3} tackles the question generation problem from Reinforcement Learning point of view. They consider generator as an actor trying to maximise BLEU score as it's reward function. \cite{qg_1} propose a Transformer based Seq2Seq pretraining model which beats the current state of the art in many summarization and question generation tasks. To the best of our knowledge, we are the first ones to propose an implication generator module to improve consistency of any VQA architecture.

\textbf{Cyclic Framework:} Cyclic training for singular modality has been used in the past for tasks such as motion tracking \cite{old_cycle1} and text-based question answering \cite{old_cycle2}. For multi-modal tasks such as VQA, cyclic training was first introduced by \cite{cycle}. They used a Visual Question Generator (VQG) module to generate rephrasings of the original question and then trained their VQA module on those rephrasings in a cyclic manner. Similar to \cite{cycle}, our framework is also model-independent and can be used for any VQA architecture. However, their aim was to make VQA models more robust to linguistic variations through rephrasings. Our aim, through our approach, is to make the models more accurate to not just rephrasings like in \cite{cycle}, but also on implications.

\section{Approach}

We use the rule-based approach in \cite{imps} to generate implications on entire VQA v2.0 dataset, referred to as the rule-based implication dataset. This rule-based method is unable to create all 3 implications for every QA pair, especially on yes/no type questions. Due to these restrictions by this approach, the rule-based implication dataset contains implications from about 60\% of the original dataset. Moreover, all generated implications are of 'yes/no' type, this serves as a strong prior for our implication generator module. Additional details about the rule-based implication dataset can be found in Section \ref{sec:exp}.

\subsection{Implication Generator Module}
The role of this module is to generate implications of a given QA pair. This can be formulated as a transformation $G : (Q,A) \longrightarrow Q_{imp}$ where $Q_{imp}$ is the generated implication. In the VQA setting, this QA pair is provided by the VQA model. Any generic VQA model takes $(Q,I)$ to predict $A'$ where $Q$ is the original question, $I$ is the image and $A'$ is the predicted answer. Our implication generator takes as input, the learned question encoding of the original question $Q$, the predicted answer scores $A'$ and a control knob (as one hot vector) to select between the three implication categories.

The implication generation module consists of three linear encoders that transform question encoding obtained from VQA model, the predicted answer scores, and the knob to lower dimensional feature vectors. These three inputs are then added together, and passed through a single layered LSTM with hidden size of 1024. This LSTM is trained to generate implications and optimized by minimizing the negative log likelihood with corresponding ground truth implication from the rule-based implication dataset. One thing to note is that we use answers scores over the entire vocabulary instead of a particular answer label, which increases performance on questions with more than one possible correct answer. Also, this provides a distribution over the entire set of answers which is a slightly rich and dense signal to learn from. 

{\renewcommand{\arraystretch}{1.2}
\begin{table*}[t]
\begin{center}
\begin{tabular}{|l c|c c c c|c|c|}
\hline
\multirow{2}{*}{\textbf{Method}} & \multirow{2}{*}{\textbf{Val acc}} & \multicolumn{4}{c|}{\textbf{Consistency(rule-based)}} &
\multirow{2}{*}{\parbox{2cm}{ \centering \textbf{Consistency} \\ \textbf{(VQA-Imp)}}} & \multirow{2}{*}{\textbf{Robustness}}\\
& & Logeq & Nec & Mutex & Overall & &\\
\hline
BUTD \cite{butd} & 63.62 & 64.3 & 71.1 & 59.8 & 65.3 & 67.14 & 79.21\\
BUTD + IQ & 63.57 & \textbf{88.5} & \textbf{96.7} & \textbf{77.0} & \textbf{88.1} & \textbf{74.38} & \textbf{80.77}\\
\hline
BAN \cite{BAN} & 65.37 & 67.1 & 77.6 & 61.1 & 69.0 & 66.57 & 79.93\\
BAN + IQ & 65.28 & \textbf{89.3} & \textbf{97.9} & \textbf{79.8} & \textbf{89.6} & \textbf{74.61} & \textbf{81.62}\\
\hline
Pythia \cite{pythia} & 64.70 & 69.7 & 76.4 & 67.7 & 70.0 & 70.89 & 79.31\\
Pythia + IQ & 65.60 & \textbf{88.7} & \textbf{97.6} & \textbf{79.0} & \textbf{88.7} & \textbf{76.55} & \textbf{82.40}\\
\hline
\end{tabular}
\end{center}
\caption{\textbf{Consistency and robustness performance on rule-based validation, VQA-Implications and VQA-Rephrasings dataset.} Consistency and robustness are defined as percentage of correctly answered implications and rephrasings respectively, generated only on correctly answered original questions. All the models trained with our approach outperform their respective baselines in all categories, keeping the validation accuracy almost same.}
\label{tab:cons}
\end{table*}}

The implication generator module - by generating implications - introduces stronger linguistic variations than rephrasings as proposed in \cite{cycle}. Thus we believe that by learning on these implications, models trained with our approach should also perform better on rephrasings thus leading to improvement in robustness, in addition to consistency.

\subsection{Knob Mechanism}

Instead of using an implied answer selected randomly from $(yes, no)$ as input to the implication generator module, we use a three way knob to switch between logical equivalence, necessary condition and mutual exclusion. This helps the model to have better control over the generated implications. In our training dataset, implications from two categories - logical equivalence and necessary condition have 'yes' as the correct answer. While training the implication generator using implied answer, we noted that model tends to generate necessary implications when provided 'yes' as the implied answer. We believe that generating a necessary condition is easier as compared to logical equivalence and without having any control signal, model might learn to generate necessary implications all the time. Hence, we provide this control signal in the form of a one hot vector between the three implication categories.

\subsection{Cyclic Framework}
To integrate our implication generator module with any VQA module, we use a cyclic framework. The confidence score over answers generated by the VQA module is used by the implication generator module. The implications are then passed as question to the VQA module, along with the image $I$ to give implied answer $A_{imp}$. This enables the VQA module to learn on these implications and improve its consistency. Inspired by \cite{cycle}, We incorporate gating mechanism and late activation in our cyclic architecture. So, instead of passing all implied questions, we filter out undesirable implications which have cosine similarity less than threshold $T_{sim}$ with the ground truth implication. Also, as part of the late activation scheme, we disable cyclic training before $A_{iter}$.

We use three loss functions in our architecture, namely VQA loss $L_{vqa}$, question loss $L_Q$ and implication loss $L_{imp}$. $L_{vqa}$ is the standard binary cross-entropy (BCE) loss between predicted answer $A'$ and ground truth $A^{gt}$. $L_Q$ is the negative log likelihood loss between generated implication $Q_{imp}$ and ground truth implication $Q_{imp}^{gt}$. $L_{imp}$ is also the BCE loss between $A_{imp}$ and $A_{imp}^{gt}$. Combining the three losses with their respective weights, we get total loss $L_{tot}$ as:

\begin{multline}
L_{tot} = L_{vqa}(A', A^{gt}) + \lambda_Q L_Q(Q_{imp}, Q_{imp}^{gt})\\ + \lambda_{imp} L_{imp}(A_{imp}, A_{imp}^{gt})   
\end{multline}

\begin{figure*}[t]
    \centering
        \includegraphics[width=\textwidth]{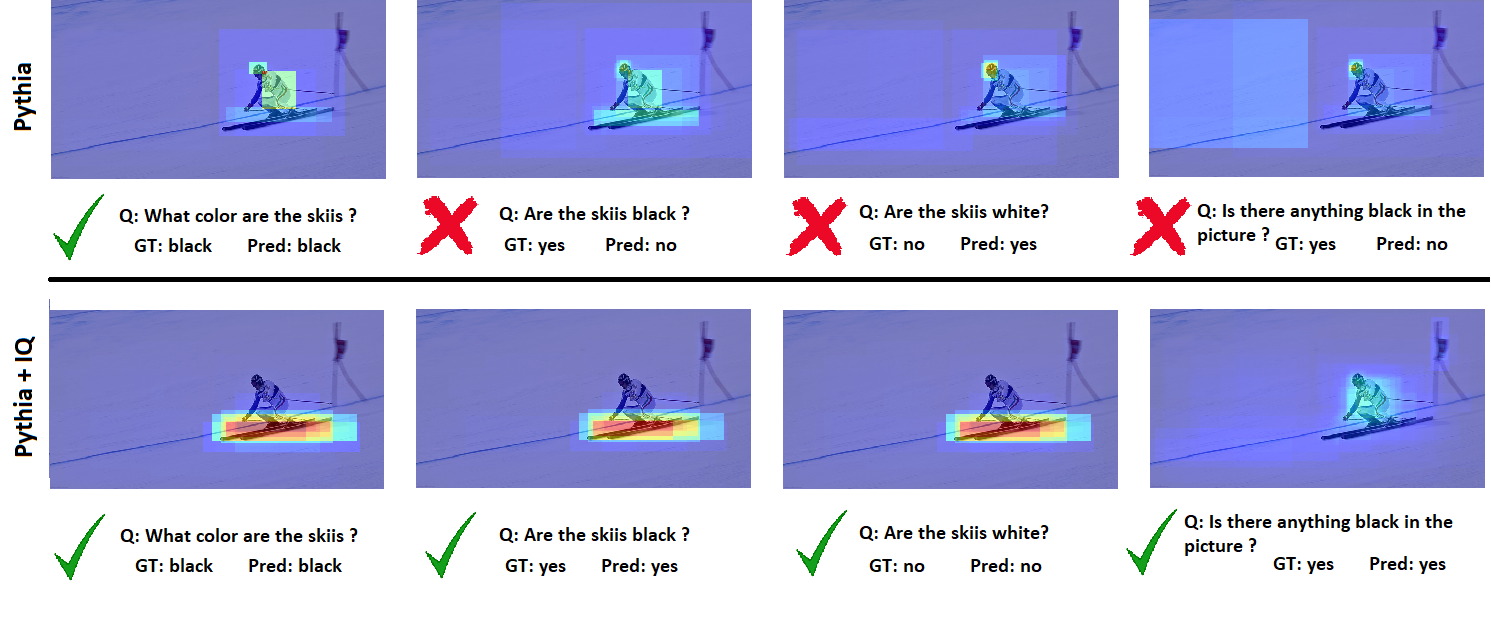}
        \includegraphics[width=\textwidth]{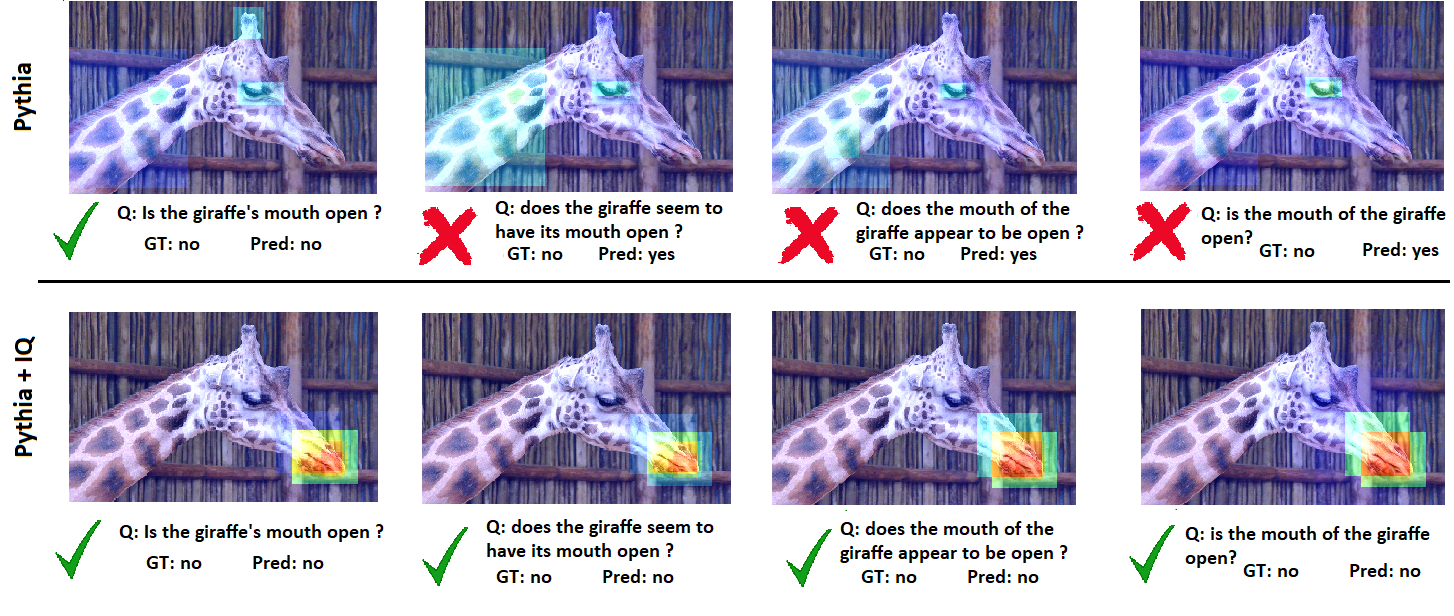}
        \caption{\textbf{Qualitative comparison of attention maps for Pythia \cite{pythia} and Pythia + IQ}. Top 2 rows represent implications and bottom 2 rows represent rephrasings. As evident from the figure, Pythia does not attend to relevant regions, whereas upon using our approach, the attention maps are much more focused on relevant regions.} 
        \label{fig:main_att}
\end{figure*}

\section{Experiments Setup} \label{sec:exp}
\subsection{Datasets}
We use the VQA v2.0 dataset for training and evaluating our model's VQA performance. The VQA v2.0 training split consists of 443,757 questions on 82,783 images and the validation split contains 214,354 questions over 40,504 images.

To train and evaluate our implication generator module and consistency, we use the implication dataset made by the rule-based approach in \cite{imps}. This dataset consists of 531,091 implied questions in training split and 255,682 questions for the validation split.

We also evaluate our model's consistency performance on human annotated VQA-Implications dataset which consists of 30,963 questions. For this dataset, we randomly select 10,500 questions from the VQA v2.0 validation set and create 3 implications (logeq, nec and mutex) per question. 

For robustness evaluation, we use the VQA-Rephrasings dataset provided by \cite{cycle}. The dataset consists of 121,512 questions by making 3 rephrasings from 40,504 questions on the VQA v2.0 validation set.

\subsection{VQA Models}
In order to show the model independent behaviour of our proposed method, we evaluate intelligence of three VQA models: BUTD, BAN, Pythia. We use the open-source implementation of these models for training and evaluation. These models are trained with hyperparameters proposed in respective papers. 

\textbf{BUTD \cite{butd}} uses bottom up attention mechanism from pretrained Faster-RCNN features on the images. Visual Genome \cite{vg} dataset is used to pretrain and extract top-K objects in the images during the preprocessing step. This model won the annual VQA challenge in 2017. For training BUTD, we used the fixed top-36 objects RCNN features for every image. Their model achieves 63.62\% accuracy on the VQA v2.0 validation split.

\textbf{BAN \cite{BAN}} uses bilinear model to reduce the computational cost of learning attention distributions, whereby different attention maps are built for each modality. Further, low-rank bilinear pooling extracts the joint representations for each pair of channels. BAN achieves 65.37\% accuracy on the VQA v2.0 validation split.

\textbf{Pythia \cite{pythia}} extracts image features from detectron also pretrained over visual genome. It also uses Resnet-152 features and ensembling over 30 models, but we didn't use these techniques in our study. Glove embeddings are used for question and its implications. Pythia was the winning entry of 2018 VQA challenge and achieves 64.70\% accuracy on the VQA v2.0 validation split.

\subsection{Implementation details}
For the gating mechanism and late activation, $T_{sim}=0.9$ and $A_{iter}=5500$ for Pythia and $A_{iter}=10,000$ for BAN and BUTD. The LSTM hidden state size for implication generator module is 1024 and Glove embeddings are used of $dim=300$. The weights for the losses are kept as $\lambda_Q=0.5$ and $\lambda_{imp}=1.5$. All models are trained on training split and evaluated on validation split of VQA v2.0 dataset. 

\section{Results and Analysis}

\subsection{Consistency performance}
As defined in Section \ref{sec:intro}, consistency of any VQA model is its ability to answer the implications of a question correctly, if it correctly answers the original question. Implications are generated on the correctly answered questions from validation VQA v2.0 dataset, and consistency score is calculated as the fraction of correct predictions to total implications. These generated implications are binary yes/no questions, and hence randomly answering them would give about 50\% consistency score.

As seen in Table \ref{tab:cons} All the 3 models achieve an average consistency score of \httilde{70\%}. i.e. they fail 30\% of the times on implications of correctly predicted questions. Intuitively, Nec-implication serves as the neccessary condition which the models should know in order to answer the question. For eg: In order to answer "How many birds are there?", they should understand if "Are there any birds in the picture?" Consistency score of \httilde{75\%} Nec-implication shows the lack of image understanding in these models. Using our approach, the three models achieve \httilde{97\%} on Nec-implication.

{\renewcommand{\arraystretch}{1.5}
\begin{table*}[t]
\begin{center}
\begin{tabular}{|l|c|c|c|c|c|c|c|}
\hline
\textbf{Method} & \textbf{BLEU-1} & \textbf{BLEU-2} & \textbf{BLEU-3} & \textbf{BLEU-4} & \textbf{ROUGE-L} & \textbf{METEOR} & \textbf{CIDEr} \\
\hline
Pythia + IQ & 0.627 & 0.520 & 0.443 & 0.381 & 0.632 & 0.288 & 3.343 \\
\hline
Pythia + IQ + Knob & \textbf{0.785} & \textbf{0.715} & \textbf{0.647} & \textbf{0.581} & \textbf{0.795} & \textbf{0.409} & \textbf{5.263} \\
\hline
\end{tabular}
\end{center}
\caption{\textbf{Implication generation performance on rule-based Implication validation dataset.} Note that using the knob mechanism instead of an implied answer gives significant improvement.}
\label{tab:ig}
\end{table*}}
\subsection{Robustness Performance}


We evaluate our framework's robustness performance on the VQA-Rephrasings dataset introduced in \cite{cycle}. Robustness is evaluated only on correctly answered original questions. Note that just like the models in \cite{cycle}, we also do not train our models on the VQA-Rephrasings dataset. The results in Table \ref{tab:cons} show that models trained using our approach are more robust compared to baseline. This is consistent with the hypotheses that our models learn to improve on a stronger linguistic variation than rephrasings by learning on implications and hence improvement in robustness is expected.

\subsection{Attention Map Analysis}

{\renewcommand{\arraystretch}{1.2}
\begin{table}[t]
\begin{center}
\begin{tabular}{|l c c|}
\hline
\multirow{2}{*}{\textbf{Method}} & \textbf{Logeq} & \textbf{Rephrasing} \\
& \textbf{($\times 10^{-4}$)} & \textbf{($\times 10^{-4}$)} \\
\hline
BUTD \cite{butd} & 31.72 & 15.51\\
BUTD + IQ \small{(ours)} & \textbf{26.73} & \textbf{13.88}\\
\hline
BAN \cite{BAN} & 8.09 & 5.03\\
BAN + IQ \small{(ours)} & \textbf{5.41} & \textbf{3.64}\\
\hline
Pythia \cite{pythia} & 11.41 & 5.40\\
Pythia + IQ \small{(ours)} & \textbf{5.83} & \textbf{3.37}\\
\hline
\end{tabular}
\end{center}
\caption{\textbf{Attention map analysis.} Logeq (Rephrasing) denotes the mean Euclidean distance between attention maps of original question and Logeq (Rephrasing). Models trained with our approach produce better results highlighting stronger multi-modal understanding.}
\label{tab:ama}
\end{table}}

As a qualitative analysis, we compare the attention maps of \cite{pythia} with our approach. As we can see in Fig \ref{fig:main_att}, the attention maps generated by our approach are significantly better than those of \cite{pythia} for both implications and rephrasings.

To ensure appropriate visual grounding, we believe that the model should look at same regions as original question for logically equivalent and rephrased questions. As a quantitative comparison, we compute the mean Euclidean distance between attention weights for logically equivalent (Logeq) and rephrased questions with their respective original question. Table \ref{tab:ama} shows that models trained with our approach tend to focus on same regions to answer the original question, its rephrasing and its logical equivalent counterpart. These analysis show that multi-modal understanding of vision and language is enhanced using our approach.

\subsection{Data Augmentation}
Since we are using an extra dataset (rule-based implications) in addition to VQA v2.0 to train our models, we also compare our models' consistency with models finetuned using data augmentation. Table \ref{tab:cons_da} summarizes these results. Better performance of our models on the human annotated VQA-Implications dataset shows that models trained with our approach generalize better and hence would do better than data augmentation in the outside world.

{\renewcommand{\arraystretch}{1.2}
\begin{table}[t]
\begin{center}
\begin{tabular}{|l c c|}
\hline
\rule{0pt}{15pt}\centering \textbf{Method} & \pbox{20cm}{\textbf{Consistency} \\[0pt] \textbf{(rule-based)}} & \pbox{20cm}{\textbf{Consistency} \\[0pt] \textbf{(VQA-Imp)}}\\
\hline
BUTD + DA & \textbf{93.1} & 74.24\\
BUTD + IQ \small{(ours)} & 88.1 & \textbf{74.38}\\
\hline
BAN + DA & 87.6 & 74.33\\
BAN + IQ \small{(ours)} & \textbf{89.6} & \textbf{74.61}\\
\hline
Pythia + DA & \textbf{89.7} & 76.19\\
Pythia + IQ \small{(ours)} & 88.7 & \textbf{76.55}\\
\hline
\end{tabular}
\end{center}
\caption{\textbf{Consistency comparison of data augmentation vs our approach.} VQA-Imp denotes our VQA-Implications dataset and DA stands for models finetuned on rule-based training implications. Even though our models lack on rule-based dataset, they consistently outperform their respective baselines on the VQA-Implication dataset.}
\label{tab:cons_da}
\end{table}}

\subsection{Implication Generator Performance}

We train our implication generator on the rule-based training dataset and evaluate our module on rule-based validation split. We use common question generator metrics such as BLEU \cite{bleu}, ROUGE-L \cite{rouge}, METEOR \cite{meteor} and CIDEr \cite{cider} scores for evaluation. We also demonstrate the importance of using the knob mechanism instead of an implied answer as input to the module. Table \ref{tab:ig} shows the results of the implication generator module.

\section{Conclusion and Future Works}
Our contributions in this paper are three fold. First, we propose a model-independent cyclic training scheme for improving consistency and robustness of VQA models without degrading their performance. Second, a novel implication generator module for making implications using the question answer pair and a knob mechanism. Third, a new annotated VQA-Implications dataset as an evaluation baseline for future works in consistency.

Our implication generator being trained on rule-based implications dataset, has its own limitations. Firstly, the implications are restricted to 3 types - Logical Equivalence, Necessary Condition and Mutual Exclusion and all implications are limited to 'yes/no' type. We believe that learning on implications not restricted to these limitations should lead to better performance. Furthermore, the rule-based implications come from a fixed distribution and are not as diverse as human annotated implications would be. This limitation can also be quantitatively seen by observing the difference between models' performance on rule-based and human annotated implications.  

\section*{Acknowledgments}
We would like to thank Manish Borthakur, Aditi Jain and Udita Mittal from Indian Institute of Technology, Delhi for annotating the VQA-Implications dataset.

{\small
\bibliographystyle{ieee_fullname}
\bibliography{egbib}
}

\begin{figure*}[p]
    \centering
    \includegraphics[width=.95\linewidth , height = .27\paperheight]{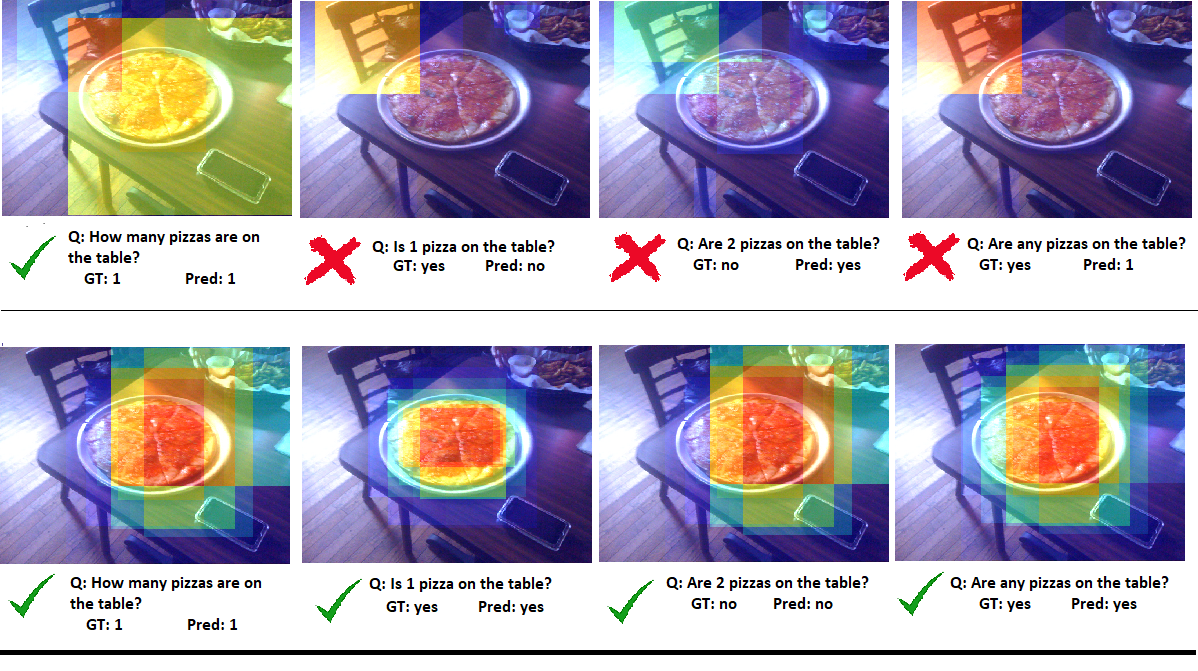}
    \includegraphics[width=.95\linewidth , height = .27\paperheight]{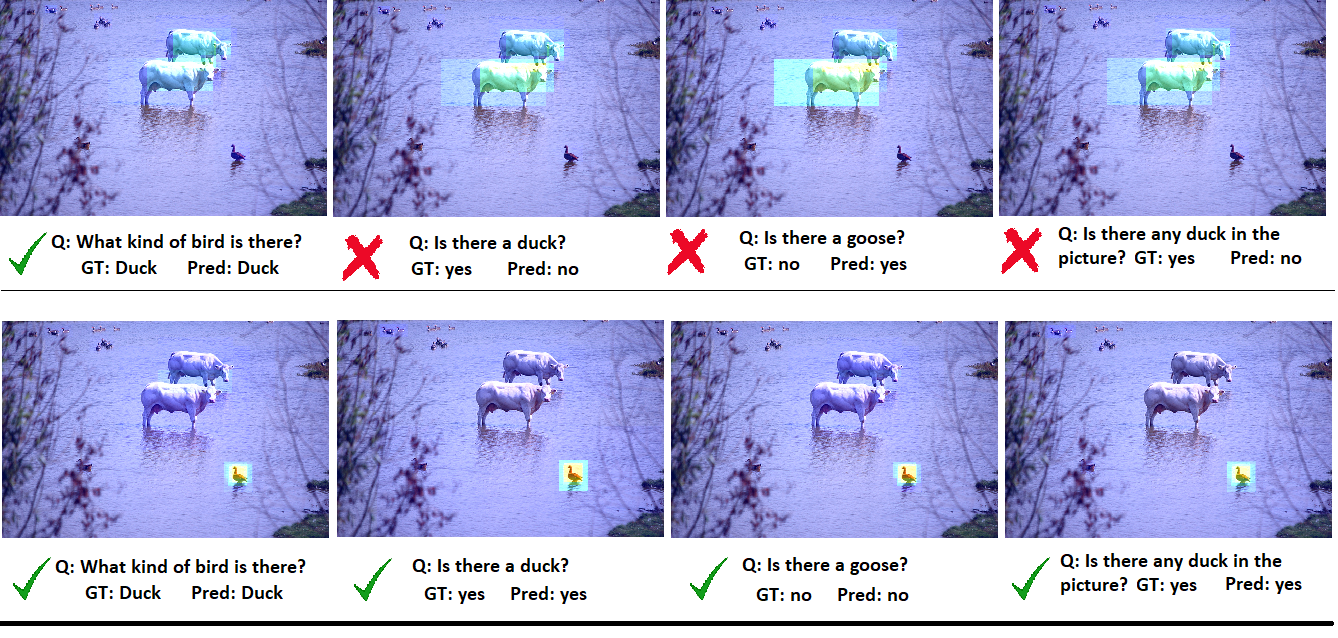}
    \includegraphics[width=.95\linewidth , height = .264\paperheight]{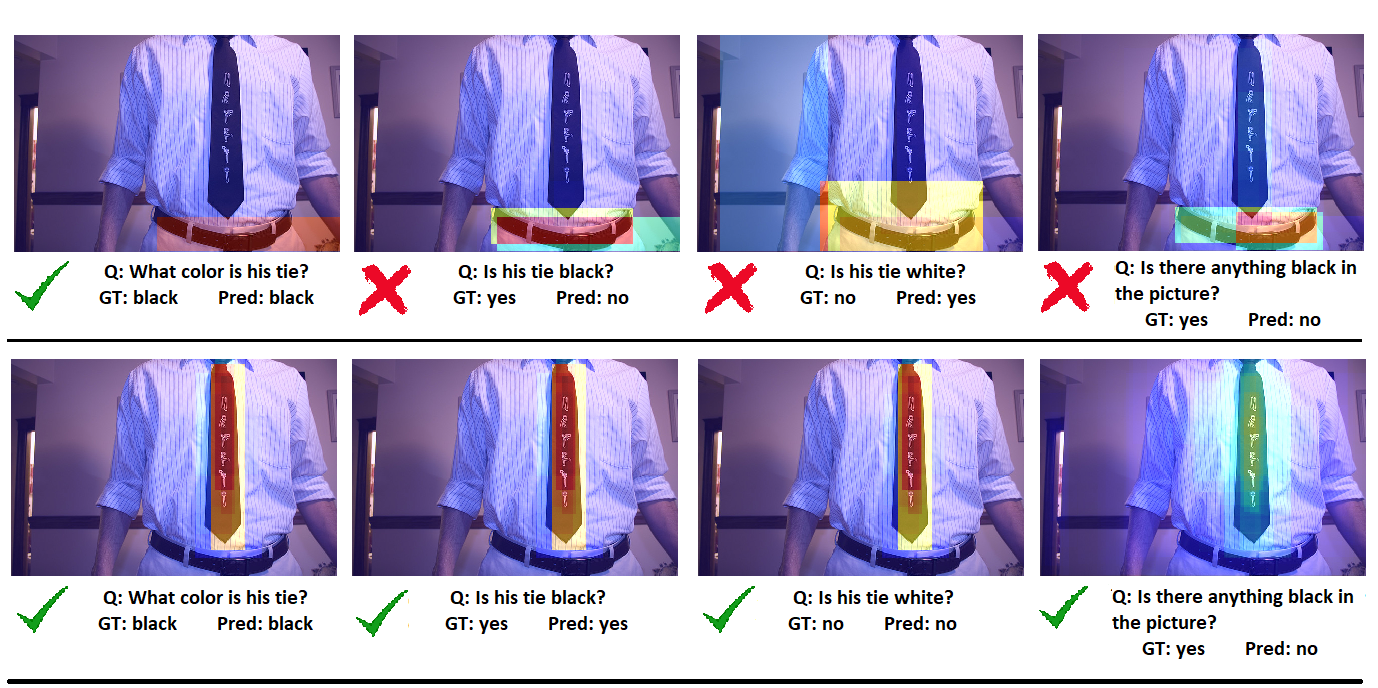}
\end{figure*}

\begin{figure*}[p]
    \centering
    \includegraphics[width=.95\linewidth , height = .27\paperheight]{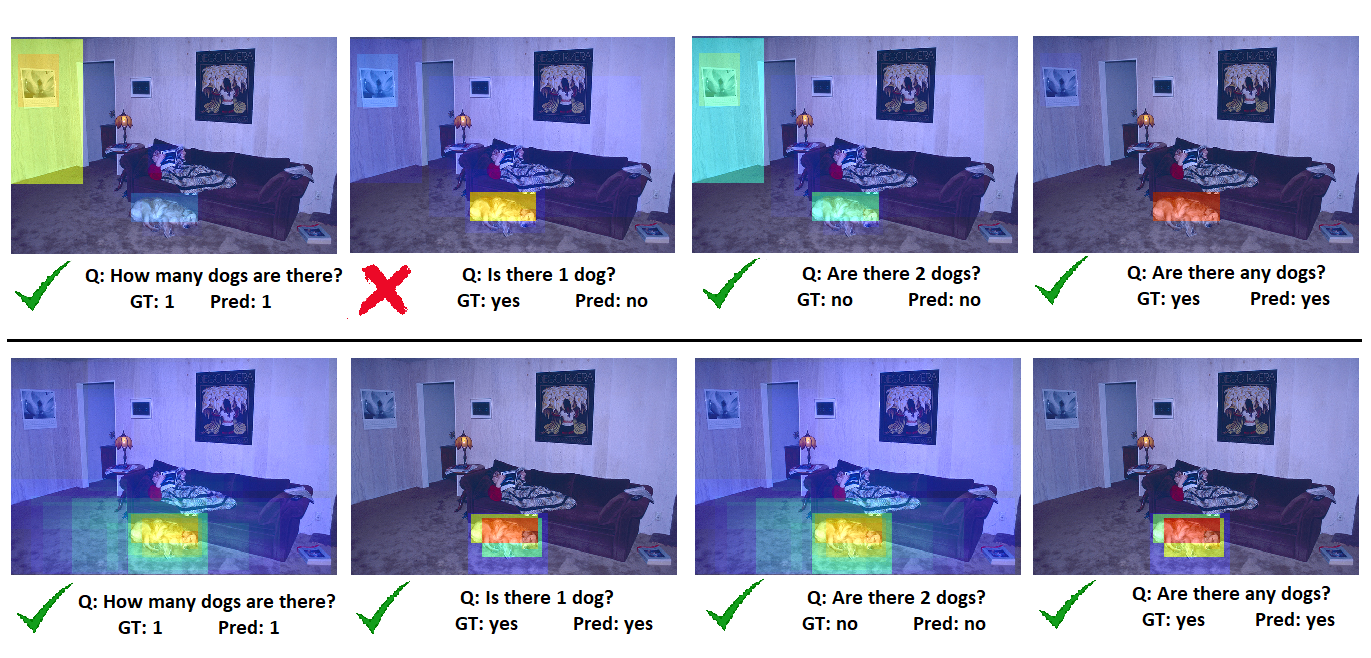}
    \includegraphics[width=.95\linewidth , height = .27\paperheight]{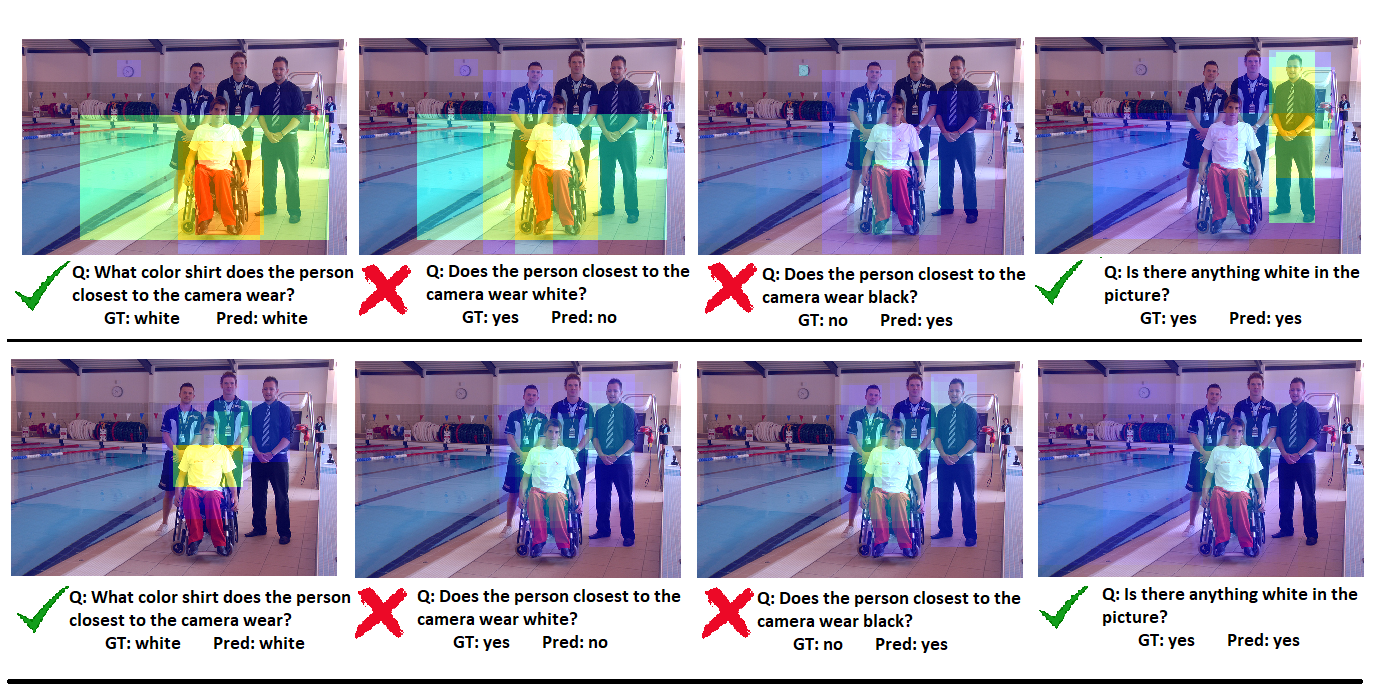}
     \caption{Attention map comparisons of Pythia \cite{pythia} vs the same model trained with our approach.}
\end{figure*}

\newpage
\begin{figure*}[p]
    \centering
    \includegraphics[scale=0.7]{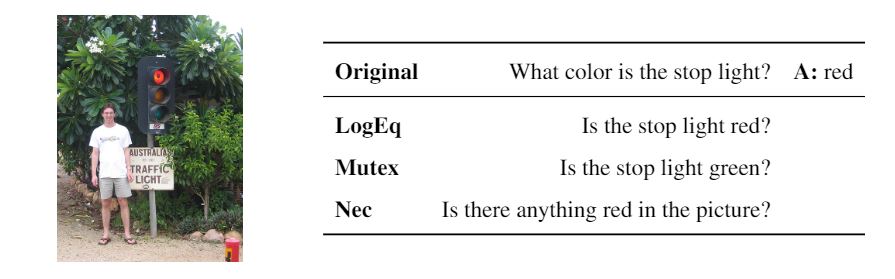}
    \includegraphics[scale=0.7]{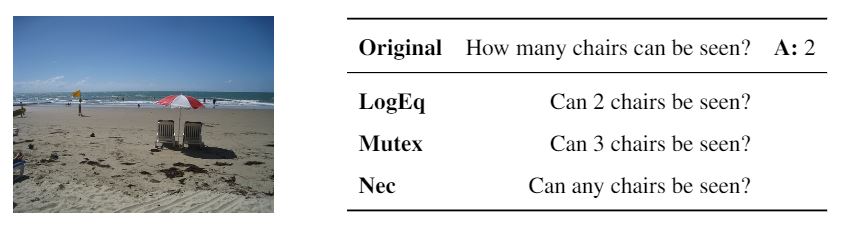}
    \includegraphics[scale=0.7]{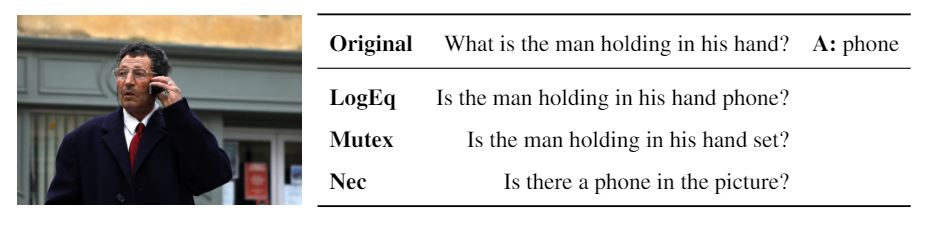}
    \includegraphics[scale=0.7]{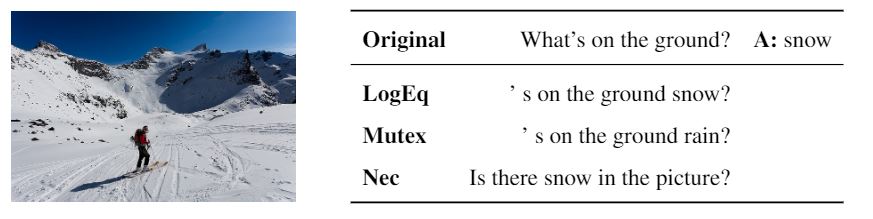}
\end{figure*}

\begin{figure*}[p]
    \centering
    \includegraphics[scale=0.7]{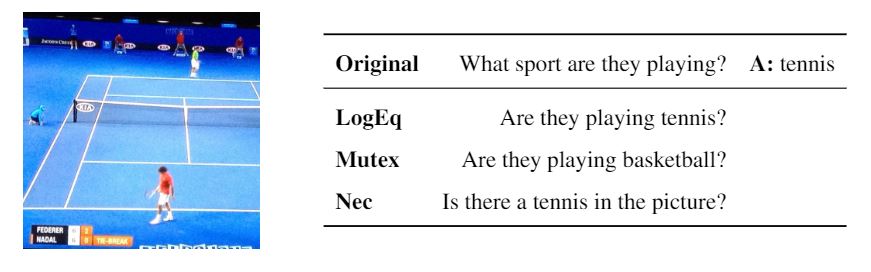}
    \includegraphics[scale=0.7]{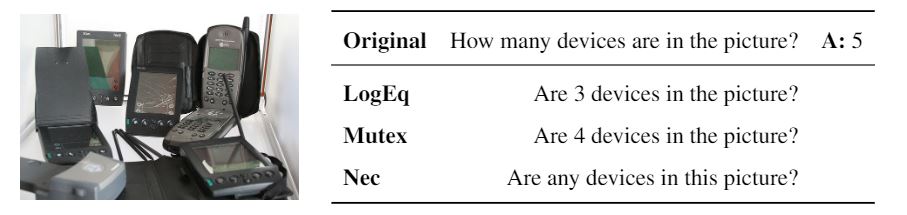}
    \includegraphics[scale=0.7]{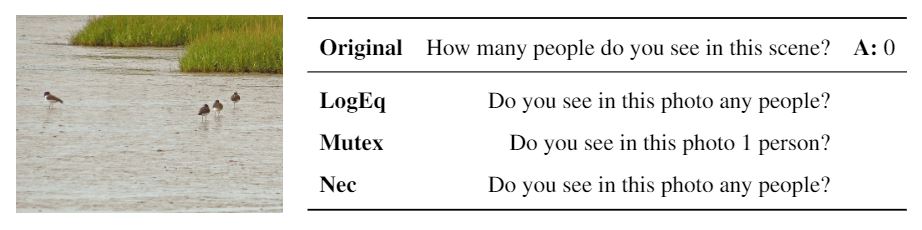}
    \includegraphics[scale=0.7]{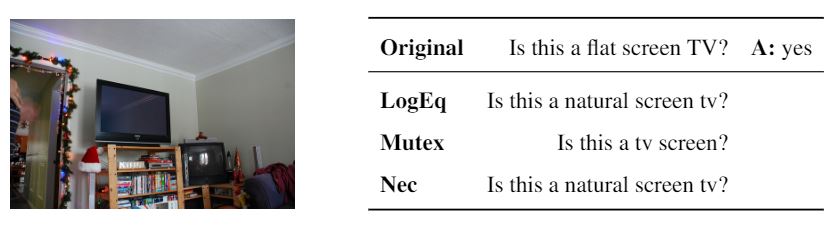}
    \caption{\textbf{Implications generated by our module.} As seen in the examples, the module can replace the answer value in Logical Equivalence type sometimes. Also, for numbered questions having answer '0' and 'yes/no' questions, the module fails to generate correct implications due to limitations of the rule-based dataset.}
\end{figure*}

\end{document}